\pdfoutput = 1
\documentclass{article}
\usepackage{mathptmx}
\usepackage[T1]{fontenc}
\usepackage[utf8]{inputenc}
\usepackage{url}
\usepackage{amssymb}
\usepackage{amsmath}
\usepackage{graphicx}
\usepackage[authoryear]{natbib}
\usepackage[unicode=true,pdfusetitle,
 bookmarks=true,bookmarksnumbered=false,bookmarksopen=false,
 breaklinks=false,pdfborder={0 0 0},backref=false,colorlinks=false]
 {hyperref}

\makeatletter
\usepackage{sidecap} 

\usepackage{graphicx} 

\usepackage[caption=false,singlelinecheck=off, position=top, farskip=0pt, nearskip=0pt, captionskip=0pt, topadjust=2pt]{subfig}
\usepackage{natbib}

\usepackage{algorithm}
\usepackage{algorithmic}

\usepackage{hyperref}

\renewcommand\[{\begin{equation}}
\renewcommand\]{\end{equation}} 

\renewenvironment{eqnarray*}%
{\begin{eqnarray}}%
{\end{eqnarray}}

\usepackage[accepted]{icml2013}

\renewcommand{\ICML@appearing}{Submitted to ICLR 2014.}

\@ifundefined{showcaptionsetup}{}{%
 \PassOptionsToPackage{caption=false}{subfig}}
\usepackage{subfig}
\makeatother

\begin{document}
\twocolumn[ \icmltitle{Neuronal Synchrony in Complex-Valued Deep Networks}
\icmlauthor{David P.~Reichert}{david\_reichert@brown.edu} 
\icmlauthor{Thomas Serre}{thomas\_serre@brown.edu}
\icmladdress{Department of Cognitive, Linguistic \& Psychological Sciences, Brown University}
\icmladdress{\emph{Appearing in the proceedings of the 2nd International Conference on Learning Representations (ICLR2014).}}
\vskip 0.3in ]
\begin{abstract}
Deep learning has recently led to great successes in tasks such as
image recognition \citep[e.g][]{krizhevsky_imagenet_2012-1}. However,
deep networks are still outmatched by the power and versatility of
the brain, perhaps in part due to the richer neuronal computations
available to cortical circuits. The challenge is to identify which
neuronal mechanisms are relevant, and to find suitable abstractions
to model them. Here, we show how aspects of spike timing, long hypothesized
to play a crucial role in cortical information processing, could be
incorporated into deep networks to build richer, versatile representations.

We introduce a neural network formulation based on complex-valued
neuronal units that is not only biologically meaningful but also amenable
to a variety of deep learning frameworks. Here, units are attributed
both a firing rate and a phase, the latter indicating properties of
spike timing. We show how this formulation qualitatively captures
several aspects thought to be related to neuronal synchrony, including
gating of information processing and dynamic binding of distributed
object representations. Focusing on the latter, we demonstrate the
potential of the approach in several simple experiments. Thus, neuronal
synchrony could be a flexible mechanism that fulfills multiple functional
roles in deep networks. \vspace{-0.11cm}

\end{abstract}

\section{Introduction}

Deep learning approaches have proven successful in various applications,
from machine vision to language processing \citep{bengio_representation_2012}.
Deep networks are often taken to be inspired by the brain as idealized
neural networks that learn representations through several stages
of non-linear processing, perhaps akin to how the mammalian cortex
adapts to represent the sensory world. These approaches are thus also
relevant to computational neuroscience \citep{cadieu_neural_2013}:
for example, convolutional networks \citep{lecun_backpropagation_1989}
possibly capture aspects of the organization of the visual cortex
and are indeed closely related to biological models like HMAX \citep{serre_feedforward_2007},
while deep Boltzmann machines \citep{salakhutdinov_deep_2009} have
been applied as models of generative cortical processing \citep{reichert_charles_2013}. 

The most impressive recent deep learning results have been achieved
in classification tasks, in a processing mode akin to rapid feed-forward
recognition in humans \citep{serre_feedforward_2007}, and required
supervised training with large amounts of labeled data. It is perhaps
less clear whether current deep networks truly support neuronal representations
and processes that naturally allow for flexible, rich reasoning about
e.g.~objects and their relations in visual scenes, and what machinery
is necessary to learn such representations from data in a mostly unsupervised
way. At the implementational level, there is a host of cortical computations
not captured by the simplified mechanisms utilized in deep networks,
from the complex laminar organization of the cortex to dendritic computations
or neuronal spikes and their timing. Such mechanisms might be key
to realizing richer representations, but the challenge is to identify
which of these mechanisms are functionally relevant and which can
be discarded as mere implementation details.

One candidate mechanism is temporal coordination of neuronal output,
or in particular, synchronization of neuronal firing. Various theories
posit that synchrony is a key element of how the cortex processes
sensory information \citep[e.g.][]{von_der_malsburg_correlation_1981,crick_function_1984,singer_visual_1995,fries_mechanism_2005,uhlhaas_neural_2009,stanley_reading_2013},
though these theories are also contested \citep[e.g.][]{shadlen_synchrony_1999,ray_differences_2010}.
Because the degree of synchrony of neuronal spikes affects the output
of downstream neurons, synchrony has been postulated to allow for
gating of information transmission between neurons or whole cortical
areas \citep{fries_mechanism_2005,benchenane_oscillations_2011}.
Moreover, the relative timing of neuronal spikes may carry information
about the sensory input and the dynamic network state \citep[e.g.][]{geman_invariance_2006,stanley_reading_2013},
beyond or in addition to what is conveyed by firing rates. In particular,
neuronal subpopulations could dynamically form synchronous groups
to \emph{bind} distributed representations \citep{singer_binding_2007},
to signal that perceptual content represented by each group forms
a coherent entity such as a visual object in a scene.

Here, we aim to demonstrate the potential functional role of neuronal
synchrony in a framework that is amenable to deep learning. Rather
than dealing with more realistic but elaborate spiking neuron models,
we thus seek a mathematical idealization that naturally extends current
deep networks while still being interpretable in the context of biological
models. To this end, we use complex-valued units, such that each neuron's
output is described by both a firing rate and a phase variable. Phase
variables across neurons represent relative timing of activity.

In Section \ref{sec:Neuronal-synchrony}, we briefly describe the
effect of synchrony on neuronal information processing. We present
the framework based on complex-valued networks, and show what functional
roles synchrony could play, within this framework. Thanks to the specific
formulation employed, we had some success with converting deep nets
trained without synchrony to incorporate synchrony. Using this approach,
in Section \ref{sec:Experiments:-the-case} we underpin our argument
with several simple experiments, focusing on binding by synchrony.
Exploiting the presented approach further will require learning with
synchrony. We discuss principled ways to do so and challenges to overcome
in Section~\ref{sec:Discussion}.

It should be noted that complex-valued neural networks are not new
\citep[e.g.][]{zemel_lending_1995,kim_approximation_2003,nitta_orthogonality_2004,fiori_nonlinear_2005,aizenberg_multilayer_2007,savitha_metacognitive_2011,hirose_nature_2011}.
However, they do not seem to have attracted much attention within
the deep learning community---perhaps because their benefits still
need to be explored further.%
\footnote{Beyond possibly applications where the data itself is naturally represented
in terms of complex numbers.%
} There are a few cases where such networks were employed with the
interpretation of neuronal synchrony, including the work of \citet{rao_unsupervised_2008},
\citet{rao_objective_2010,rao_effects_2011}, which is similar to
ours. These prior approaches will be discussed in Section \ref{sec:Discussion}.

\section{Neuronal synchrony\label{sec:Neuronal-synchrony}}

Cortical neurons communicate with electric action potentials (so-called
spikes). There is a long-standing debate in neuroscience on whether
various features of spike timing matter to neuronal information processing,
rather than just average firing rates \citep[e.g.][]{stanley_reading_2013}.
In common deep neural networks (convolutional networks, Boltzmann
machines, etc.), the output of a neuronal unit is characterized by
a single (real-valued) scalar; the state of a network and how it relates
to an interpretation of sensory input is fully determined by the joint
scalar outputs across all units. This suggests an interpretation in
terms of average, static firing rates, lacking any notion of relative
timing. Here, we consider how to incorporate such notions into deep
networks.

\begin{figure*}[t]
\subfloat[]{\includegraphics[width=0.65\textwidth]{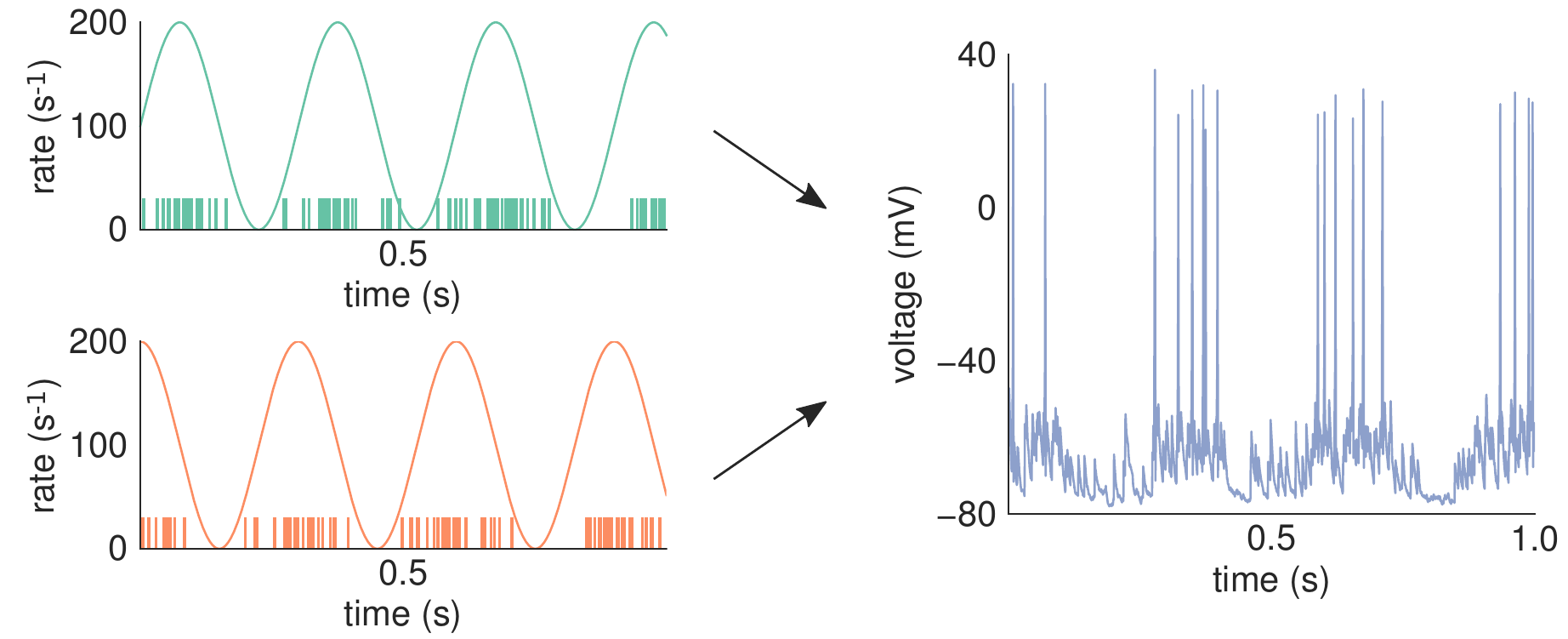}

}\hfill{}\subfloat[]{\includegraphics[width=0.3\textwidth]{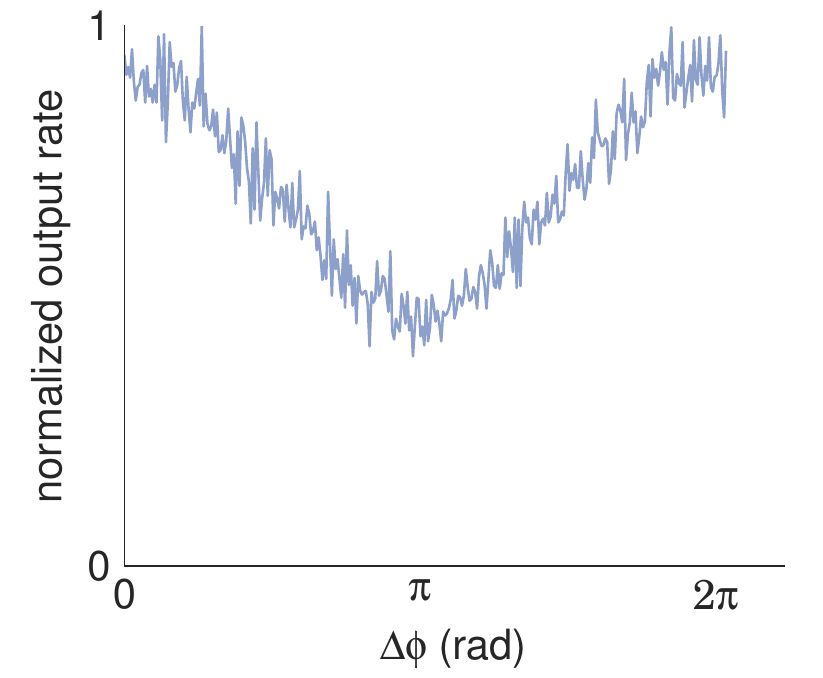}

}

\vspace{-0.1cm}
\subfloat[]{\includegraphics[width=0.3\textwidth]{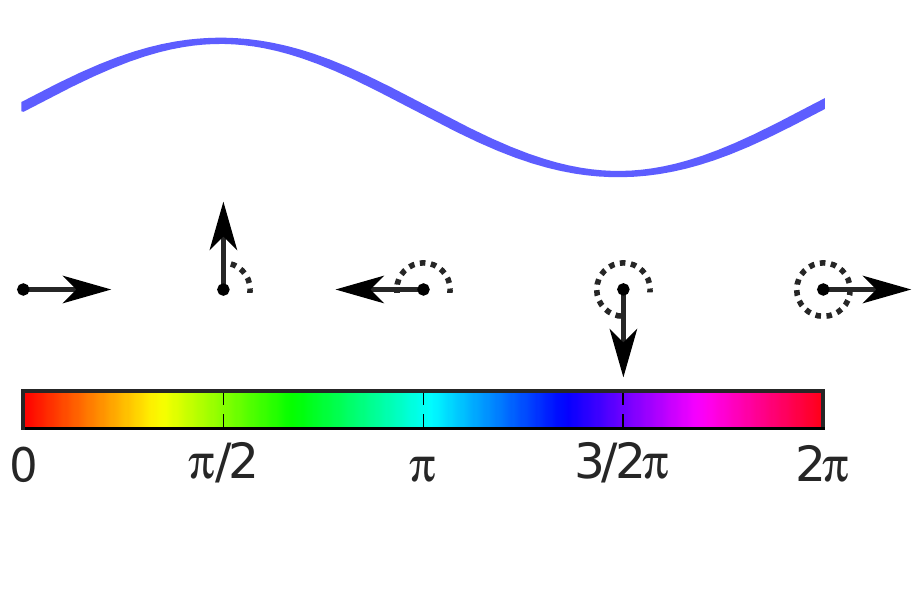}

}\hfill{}\subfloat[]{\hfill{}\includegraphics[width=0.3\textwidth]{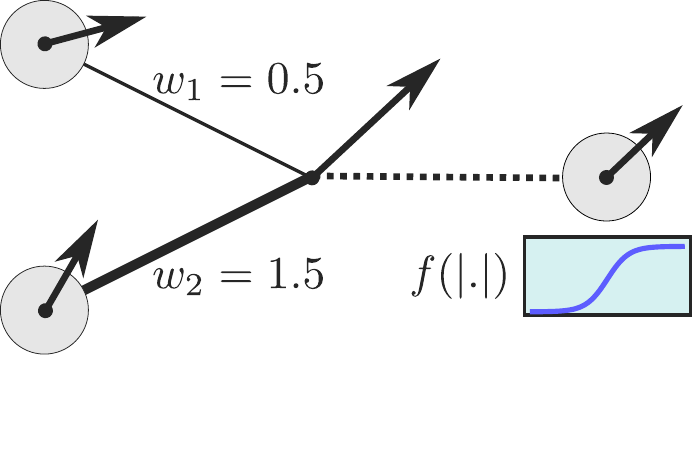}\hfill{}

}\hfill{}\subfloat[]{\includegraphics[width=0.3\textwidth]{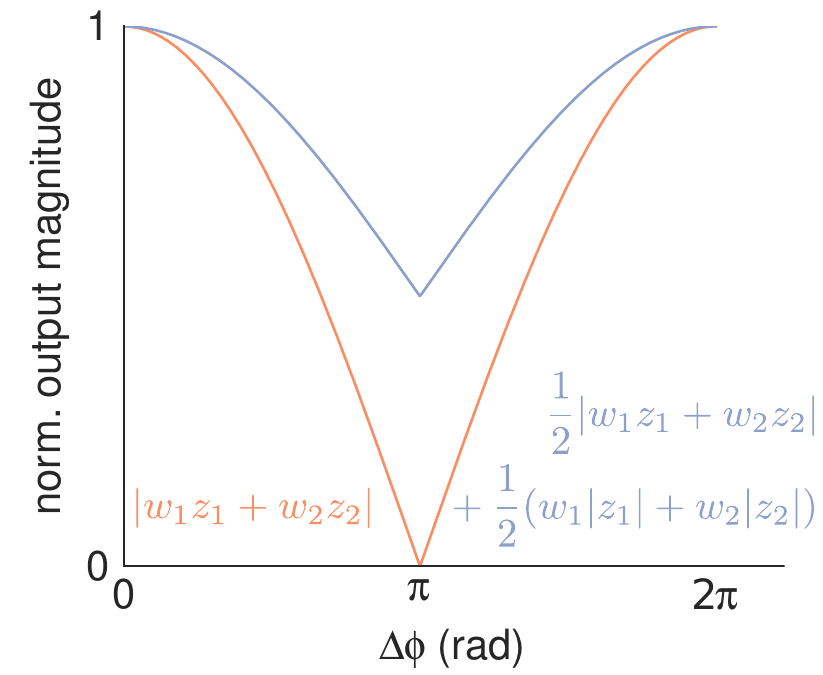}

}

\caption{\textbf{Transmission of rhythmical activity, and corresponding model
using complex-valued units.} (a) A Hodgkin–Huxley model neuron receives
two rhythmic spike trains as input, plus background activity. The
inputs are modeled as inhomogeneous Poisson processes modulated by
sinusoidal rate functions (left; shown are rates and generated spikes),
with identical frequencies but differing phases. The output of the
neuron is itself rhythmical (right; plotted is the membrane potential).
(b) The neuron's output rate is modulated by the phase difference
between the two inputs (rate averaged over 15s runs). (c) We represent
the timing of maximal activity of a neuron as the phase of a complex
number, corresponding to a direction in the complex plane. The firing
rate is the magnitude of that complex number. Also shown is the color
coding used to indicate phase throughout this paper (thus, figures
should be viewed in color). (d) The outputs of the input neurons are
scaled by synaptic weights (numbers next to edges) and added in the
complex plane. The phase of the resulting complex input determines
the phase of the output neuron. The activation function $f$ is applied
to the magnitude of the input to compute the output magnitude. Together,
this models the influence of synchronous neuronal firing on a postsynaptic
neuron. (e) Output magnitude as function of phase difference of two
inputs. With a second term added to a neuron's input, out-of-phase
excitation never cancels out completely (see main text for details;
curves are for $w_{1}=w_{2}>0$, $|z_{1}|=|z_{2}|$). Compare to 1b.\label{fig:Transmission-of-rhythmical}}
\end{figure*}

Consider Figure \ref{fig:Transmission-of-rhythmical}a for an example
of how a more dynamic code could be transmitted between neurons (simulated
with the Brian simulator, \citealp{goodman_brian_2009}). This example
is based on the hypothesis that neuronal rhythms, ubiquitous throughout
the brain, play a functional role in information processing \citep[e.g.][]{singer_visual_1995,fries_mechanism_2005,uhlhaas_neural_2009,benchenane_oscillations_2011}.
A neuron receives spike train inputs modulated by oscillatory firing
rates. This results in rhythmic output activity, with an average
firing rate that depends both on the amplitudes and relative phase
of the inputs (Figure \ref{fig:Transmission-of-rhythmical}b). Such
interactions are difficult to represent with just static firing rates.

\subsection{Modeling neuronal synchrony with complex-valued units\label{sub:Modeling-neuronal-synchrony}}

In deep networks, a neuronal unit receives inputs from other neurons
with states vector $\mathbf{x}$ via synaptic weights vector $\mathbf{w}$.
We denote the total `postsynaptic' input as $\chi:=\mathbf{w}\cdot\mathbf{x}$.
The output is computed with an activation function $f$ as $f(\chi)$
(or, in the case of Gibbs-sampling in Boltzmann machines, $f(\chi)$
is a conditional probability from which the output state is sampled).%
\footnote{Operations such as max-pooling require separate treatment. Also, bias
parameters $\mathbf{b}$ can be added to the inputs to control the
intrinsic excitability of the neurons. We omit them for brevity.%
} We can now model aspects of spike timing by replacing the real-valued
states $\mathbf{x}$ with complex states $\mathbf{z}$. For unit state
$z_{i}=r_{i}e^{\phi_{i}},$ the magnitude $r_{i}=|z_{i}|$ can be
interpreted as the average firing rate analogously to the real-valued
case. The phase $\phi_{i}$ could correspond to the phase of a neuronal
rhythm as in Figure \ref{fig:Transmission-of-rhythmical}a, or, more
generally, the timing of maximal activity in some temporal interval
(Figure \ref{fig:Transmission-of-rhythmical}c). Because neuronal
messages are now added in the complex plane (keeping the weights $\mathbf{w}$
real-valued, for now), a neuron's total input $\zeta:=\mathbf{w}\cdot\mathbf{z}$
no longer depends only on the firing rates of the input units, and
the strength of the synapses, but also their \emph{relative timing}.
This naturally accounts for the earlier, spiking neuron example: input
states that are synchronous, i.e.~have similar phases, result in
a stronger total input, whereas less synchronous inputs result in
weaker total input (Figure \ref{fig:Transmission-of-rhythmical}d).

A straightforward way to define a neuron's output state $z_{i}=r_{i}e^{i\phi_{i}}$
from the  (complex-valued) total input $\zeta$ is to apply an activation
function, $f:\mathbb{R}^{+}\mapsto\mathbb{R}^{+}$, to the input's
magnitude $|\zeta|$ to compute the output magnitude,
and to set the output phase to the phase of the total input:
\begin{equation}
\phi_{i}=\mbox{arg}(\zeta),\quad r_{i}=f(|\zeta|),\quad \mbox{where }\zeta=\mathbf{w}\cdot\mathbf{z}.\label{eq:complexOutput}
\end{equation}
 Again this is intuitive as a biological model, as the total strength
and timing of the input determine the `firing rate' and timing of
the output, respectively.

There are, however, issues with this simple approach to modeling neuronal
synchrony, which are problematic for the biological model but also,
possibly, for the functional capabilities of a network. In analogy
to the spiking neuron example, consider two inputs to a neuron that
are excitatory (i.e., $w_{1},w_{2}>0$), and furthermore of equal
magnitude, $|w_{1}z_{1}|=|w_{2}z_{2}|$. While it is desirable that
the net total input is decreased if the two inputs are out of phase,
the net input in the complex-valued formulation can actually be zero,
if the difference in input phases is $\pi$, no matter how strong
the individual inputs (Figure \ref{fig:Transmission-of-rhythmical}e,
lower curve). Biologically, it seems unrealistic that strong excitatory
input, even if not synchronized, would not excite a neuron.%
\footnote{Arguably, refractory periods or network motifs such as disynaptic
feedforward inhibition \citep{gabernet_somatosensory_2005,stanley_reading_2013}
could indeed result in destructive interference of out-of-phase excitation.%
} 

Moreover, in the above formulation, the role of inhibition (i.e.,
connections with $w<0$) has changed: inputs with negative weights
are equivalent to \emph{excitatory inputs of the opposite phase},
due to $-1=e^{i\pi}$. Again, this is a desirable property that leads
to desynchronization between neuronal groups, in line with biological
models, as we will show below. However, it also means that inputs
from connections with negative weights, on their own, can strongly
drive a neuron; in that sense, there is no longer actual inhibition
that always has a suppressive effect on neuronal outputs. Additionally, we
found that the phase shifting caused by inhibition could result in
 instability in networks with dominant negative weights, leading to fast switching of the phase variables.

We introduce a simple fix for these issues, modifying how the output
magnitude of a neuron is computed as follows:
\vspace{0.cm}
\begin{equation}
\begin{split}
r_{i}=f(|\zeta|)\hookrightarrow r_{i}=f(\frac{1}{2}|\zeta|+\frac{1}{2}\chi),\label{eq:input} \\
\mbox{where } \zeta=\mathbf{w}\cdot\mathbf{z}, \quad \chi:=\mathbf{w}\cdot|\mathbf{z}|.
\end{split}
\end{equation}
The first term, which we refer to as \emph{synchrony term}, is the
same as before. The second, \emph{classic term}, applies the weights
to the magnitudes of the input units and thus does not depend on their
phases; a network using only the classic terms reduces to its real-valued
counterpart (we thus reuse the variable $\chi$, earlier denoting
postsynaptic input in a real-valued network). Together, the presence of the classic term implies that
excitatory input always has a net excitatory component, even if the
input neurons are out of phase such that the synchrony term is zero
(thus matching the spiking neuron example,%
\footnote{Real neuronal networks and realistic simulations have many degrees
of freedom, hence we make no claim that our formulation is a general
or quantitative model of neuronal interactions.%
} compare Figures \ref{fig:Transmission-of-rhythmical}b and \ref{fig:Transmission-of-rhythmical}e).
Similarly, input from negative connections alone is never greater
than zero. Lastly, this formulation also makes it possible to give
different weightings to synchrony and classic terms, thus controlling
how much impact synchrony has on the network; we do not explore this
possibility here.

\subsection{The functional relevance of synchrony\label{sub:The-functional-relevance}}

The advantage of using complex-valued neuronal units rather than,
say, spiking neuron models is that it is natural to consider how to
apply deep learning techniques and extend existing deep learning neural
networks in this framework. Indeed, our experiments presented later
are based on pretraining standard, real-valued nets (deep Boltzmann
machines in this case) and converting them to complex-valued nets
after training. In this section, we briefly describe how our framework
lends itself to realize two functional roles of synchrony as postulated
by biological theories.

\subsubsection*{Grouping (binding) by synchrony}

The activation of a real-valued unit in an artificial neural network
can often be understood as signaling the presence of a feature or
combination of features in the data. The phase of a complex-valued
unit could provide additional information about the feature. \emph{Binding
by synchrony} theories \citep{singer_binding_2007} postulate that
neurons in the brain dynamically form synchronous assemblies to signal
where distributed representations together correspond to coherent
sensory entities. For example, different objects in a visual scene
would correspond to different synchronous assemblies in visual cortex.
In our formulation, phases can analogously signal a soft assignment
to different assemblies. 

Importantly, communication with complex-valued messages also naturally
leads to different synchronous groups emerging: for excitatory connections,
messages that `agree' \citep{zemel_lending_1995} in their phases
prevail over those that do not; inhibitory messages, on the other
hand, equate to excitatory messages of opposite phases, and thus encourage
\emph{de}synchronization between neurons. For comparison, consider
the more realistic spiking model of visual cortex of \citet{miconi_gamma_2010-1},
where synchronous groups arise from a similar interaction between
excitation and inhibition. That these interactions can indeed lead
to meaningful groupings of neuronal representations in deep networks
will be shown empirically in Section \ref{sec:Experiments:-the-case}.

\subsubsection*{Dynamic gating of interactions and information flow}

\begin{figure}
\begin{centering}
\includegraphics[width=0.45\textwidth]{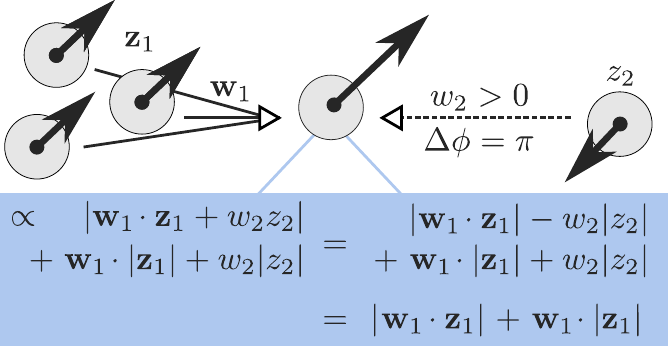}
\par\end{centering}

\caption{\textbf{Gating of interactions.} Out-of-phase input, when combined
with a stronger input, is weakened. In this example, with $\Delta\phi=\pi$
and as long as $|\mathbf{w}_{1}\cdot\mathbf{z}_{1}|>|w_{2}z_{2}|$,
effective input from the neuron to the right is zero, for any input
strength (classic and synchrony terms contributions cancel, bottom
panel). Hence, neuronal groups with different phases (gradually) decouple.\label{fig:Gating-of-interactions.}}
\end{figure}

Because synchrony affects which neuronal messages are transmitted
preferentially, it has also been postulated that synchrony may gate
information flow dynamically depending on the sensory input, the current
network state and top-down control \citep{stanley_reading_2013},
as well as to modulate the effective interactions between cortical
areas depending on their level of coherence \citep{fries_mechanism_2005,benchenane_oscillations_2011}.
A similar modulation of interactions can be reproduced in our framework.
Let us consider an example scenario (Figure \ref{fig:Gating-of-interactions.})
where a neuron is a member of a synchronous assembly, receiving excitatory
inputs $\mathbf{w}_{1}\cdot\mathbf{z}_{1}$ from neurons that all
have similar phases. Now consider the effect of adding another neuron
that also provides excitatory input, $w_{2}z_{2}$, but of a different
phase, and assume that $|\mathbf{w}_{1}\cdot\mathbf{z}_{1}|<|w_{2}z_{2}|$
(i.e.~the input of the first group dominates). The net effect the
latter additional input has depends again on the phase difference.
In particular, if the phase difference is maximal ($\pi)$, the net
contribution from the second neuron turns out to be \emph{zero}. The
output magnitude is computed as in Eq.~\ref{eq:input}, taking both
synchrony and classic terms into account. In the complex plane, $w_{2}z_{2}$
is antiparallel to $\mathbf{w}_{1}\cdot\mathbf{z}_{1}$, thus the
synchrony term is reduced by $|w_{2}z_{2}|$. However, this reduction
is exactly canceled out by the classic term contribution from the
second input (Figure \ref{fig:Gating-of-interactions.} lower panel).
There is also no effect on the output phase as the phase of the total
input remains equal to the phase of $\mathbf{w}_{1}\cdot\mathbf{z}_{1}$.
Analogous reasoning applies for inhibitory connections.

Thus, the effective connectivity between neuronal units is modulated
by the units' phases, which themselves are a result of network interactions\@.
In particular, if inference results in neurons being segregated into
different assemblies (ideally because they represent independent causes
in the sensory input, or independent regions in an image), existent
connections between groups are weakened.

\section{Experiments: the case of binding by synchrony\label{sec:Experiments:-the-case}}

In this section, we support our reasoning with several simple experiments,
and further elucidate on the possible roles of synchrony. We focus
on the binding aspect.

All experiments were based on pretraining networks as normal, real-valued
deep Boltzmann machines (DBMs, \citealp{salakhutdinov_deep_2009}).
DBMs are multi-layer networks that are framed as probabilistic (undirected
graphical) models. The \emph{visible} units make up the first layer
and are set according to data, e.g.~images. Several \emph{hidden}
layers learn internal representations of the data, from which they
can generate the latter by sampling the visible units. By definition,
in a DBM there are only (symmetric) connections between adjacent layers
and no connections within a layer. Given the inputs from adjacent
layers, a unit's state is updated stochastically with a probability
given by a sigmoid (logistic) activation function (implementing Gibbs
sampling). Training was carried out layer-wise with standard methods
including contrastive divergence (for model and training details,
see Appendix \ref{sec:Model-parameters}). Training and experiments
were implemented within the Pylearn2 framework of \citet{goodfellow_pylearn2:_2013}.

We emphasize however that our framework is not specific to DBMs, but
can in principle be adapted to various deep learning approaches (we
are currently experimenting with networks trained as autoencoders
or convolutional networks). The learning and inference procedures
of a DBM derive from its definition as a probabilistic model, but
for our purpose here it is more appropriate to simply think of a DBM
as a multi-layer recurrent neural network \citep[cf.][]{goodfellow_joint_2013}
with logistic activation function;%
\footnote{The activation function is stochastic in the case of Gibbs sampling,
deterministic in the case of mean-field inference.%
} we can demonstrate how our framework works by taking the pretrained
network, introducing complex-valued unit states and applying the activation
function to magnitudes as described in Section \ref{sub:Modeling-neuronal-synchrony}.%
\footnote{Our results were qualitatively similar whether we computed the output
magnitudes stochastically or deterministically.%
} However, developing a principled probabilistic model based on Boltzmann
machines to use with our framework is possible as well (Section \ref{sec:Discussion}). 

This conversion procedure applied to real-valued networks offers a
simple method of exploring aspects of synchrony, but there is no guarantee
that it will work (for additional discussion, see Appendix \ref{sec:Model-parameters}).
We use it here to show what the functional roles of synchrony could
be in principle; learning with synchrony will be required to move
beyond simple experiments (Section \ref{sec:Discussion}). 

Throughout the experiments, we clamped the magnitudes of the visible
units according to (binary) input images, which were not seen during
training, and let the network infer the hidden states over multiple
iterations. The phases of the visible layer were initialized randomly
and then determined by the input from the hidden layer above. Hence,
any synchronization observed was spontaneous.

\subsection{Dynamic binding of independent components in distributed representations\label{sub:Dynamic-binding-of}}

\begin{figure*}
\begin{centering}
\subfloat[]{\includegraphics[width=0.2\textwidth]{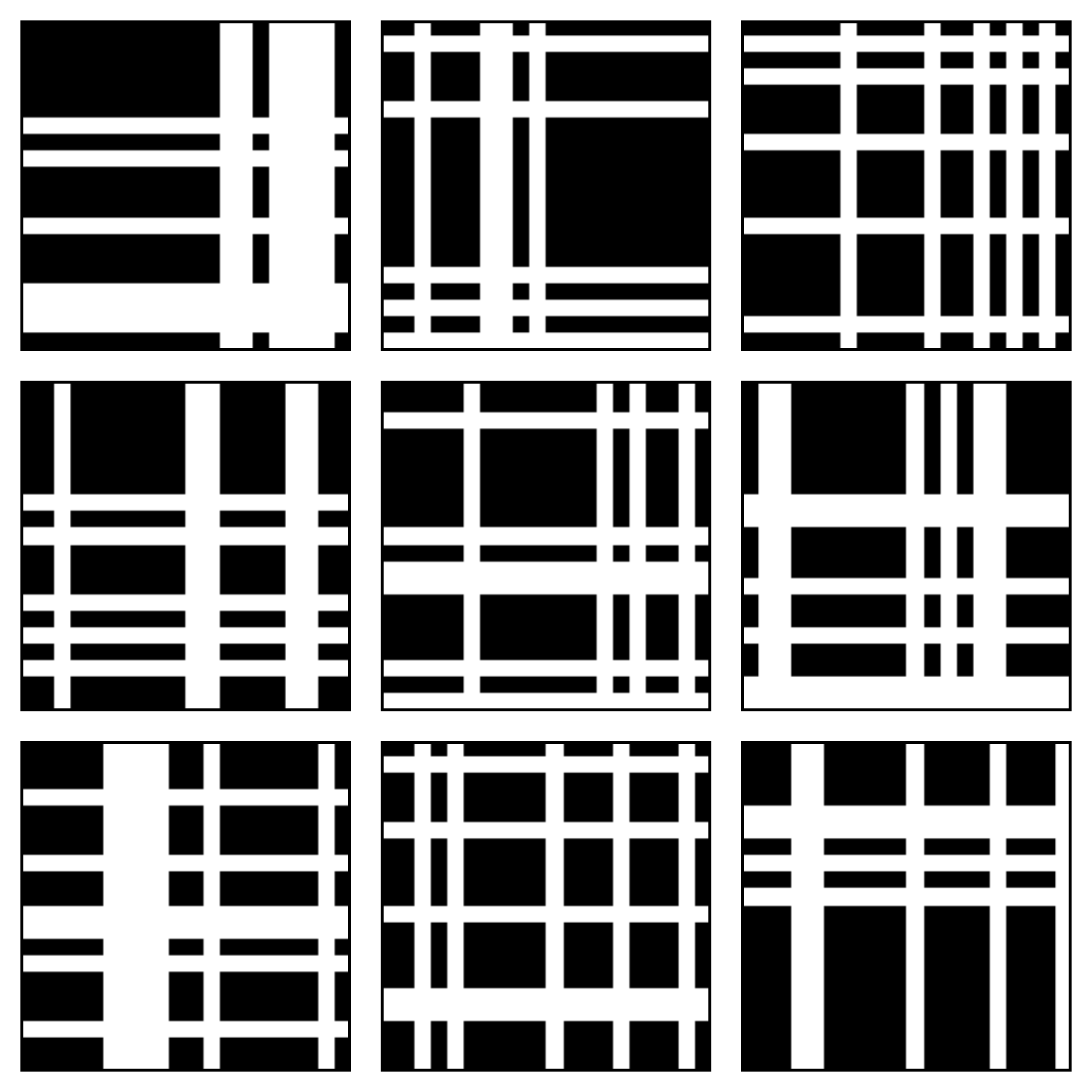}

}\hfill{}\subfloat[]{\includegraphics[width=0.35\textwidth]{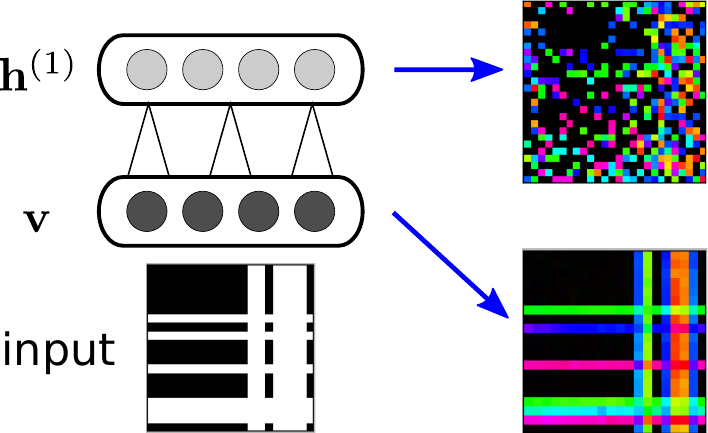}

}\hfill{}\subfloat[]{\includegraphics[width=0.25\textwidth]{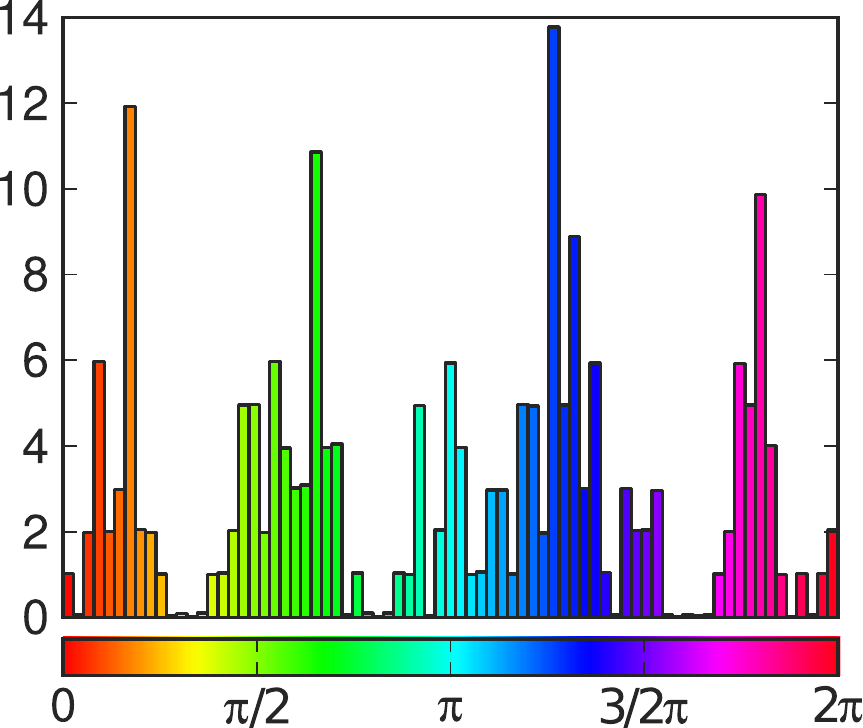}

}
\par\end{centering}

\caption{\textbf{Binding by synchrony in shallow, distributed representations.}
\label{fig:bars_results} (a) Each image of our version of the bars
problem contained 6 vertical and 6 horizontal bars at random positions.
(b) A restricted Boltzmann machine was trained on bars images and
then converted to a complex-valued network. The magnitudes of the
visible units were clamped according to the input image (bottom left),
whereas the hidden units and phases of the visible units were activated
freely. After 100 iterations, units representing the various bars
were found to have synchronized (right; the phases are color-coded
for units that are active; black means a unit is off). The neurons
synchronized even though receptive fields of the hidden units were
constrained to be smaller than the bars. Thus, binding by synchrony
could make the `independent components' of sensory data explicit
in distributed representation, in particular when no single neuron
can possibly represent a component (a full-length bar) on its own.
(c) Histogram of the unit phases in the visible layer for the example
shown in b.}
\end{figure*}

In this first experiment, we trained a DBM with one hidden layer (a
restricted Boltzmann machine, \citealp{smolensky_information_1986})
on a version of the classic `bars problem' \citep{foldiak_forming_1990},
where binary images are created by randomly drawing horizontal and
vertical bars (Figure \ref{fig:bars_results}a). This dataset has
classically been used to test whether unsupervised learning algorithms
can find the independent components that constitute the image, by
learning to represent the individual bars (though simple, the bars
problem is still occasionally employed, e.g.~\citealp{lucke_maximal_2008,spratling_unsupervised_2011}).
We chose this dataset specifically to elucidate on the role of synchrony
in the context of distributed representations.

We hard-coded the receptive field sizes (regions with non-zero weights
to the input) of the hidden units to be restricted to regions smaller
than the entire image (but together tiling the whole image). By necessity,
this implies that any individual unit can never fully represent a
full-length bar, in the sense that the the unit's weights correspond
to the bar, or that one can read out the presence of the full bar
from this unit's state alone. However, this does not imply that the
full network cannot learn that the images are constituted by bars
(as long as receptive fields overlap). For example, we found that
when sampling from the model (activating hidden and visible units
freely), the resulting images contained full-length bars most of the
time (see supplementary figure S1a and supplementary videos, Appendix
\ref{sec:Supplementary-figures}); similarly, the network would fill
in the remainder of a bar when the visible units where clamped to
a part of it.

After conversion to a complex-valued network, the model was run on
input images for 100 iterations each. Results are plotted in Figure
\ref{fig:bars_results}b, depicting both visible and hidden states
for one input image (Further examples in Figure S1b). We found that
visible neurons along a bar would often synchronize to the same phase
(except where bars crossed), whereas different bars tended to have
different phase values. Figure \ref{fig:bars_results}c shows a histogram
of phase values in the visible layer for this example image, with
clear peaks corresponding to the phases of the bars. Such synchronization
was also found in the hidden layer units (\ref{fig:bars_results}b).

Based on these results, we make three points. First, the results show
that our formulation indeed allows for neurons to dynamically organize
into meaningful synchronous assemblies, even without supervision towards
what neurons should synchronize to, e.g.~by providing phase targets
in training---here, synchrony was not used in training at all. That
the conversion from a real-valued network can work suggests that an
unsupervised or semi-supervised approach to learning with synchrony
could be successful as long as synchrony benefits the task at hand.

Second, synchronization of visible and hidden units, which together
represent individual bars, can occur for neurons several synapses
apart. At the same time, not all bars synchronized to different phases.
The number of distinct, stable phase groups that can be formed is
likely to be limited. Notably, it has been argued that this aspect
of synchrony coding explains certain capacity limits in cognition
\citep{jensen_hippocampal_2005,fell_role_2011}.

The third point relates to the nature of distributed representations.
For the bars problem, whether a neural net (or probabilistic model)
discovers the bars is usually evaluated by examining whether individual
units correspond to individual bars, as can be seen by inspecting
the weights or by probing the response properties of individual neurons
\citep[e.g.][]{lucke_maximal_2008,spratling_unsupervised_2011}. A
similar `localist' approach was taken in recent attempts to make
sense of the somewhat opaque hidden representations learned by deep
networks (as in the example of the neurons that discovered the `concept'
of a cat from unsupervised learning \citealp{le_building_2011}, or
the work of \citealp{zeiler_visualizing_2013} on analyzing convolutional
networks). In our experiment, it is not possible to map individual
neurons to the image constituents, by construction; bars could only
be represented in a distributed fashion. Synchrony could make explicit
which neurons together represent a sensory entity, e.g.~for a readout
(more on that below), as well as offer a mechanism that establishes
the grouping in the first place.

\subsection{Binding in deep networks\label{sub:Binding-in-deep}}

\begin{figure}
\subfloat[]{\includegraphics[width=0.12\textwidth]{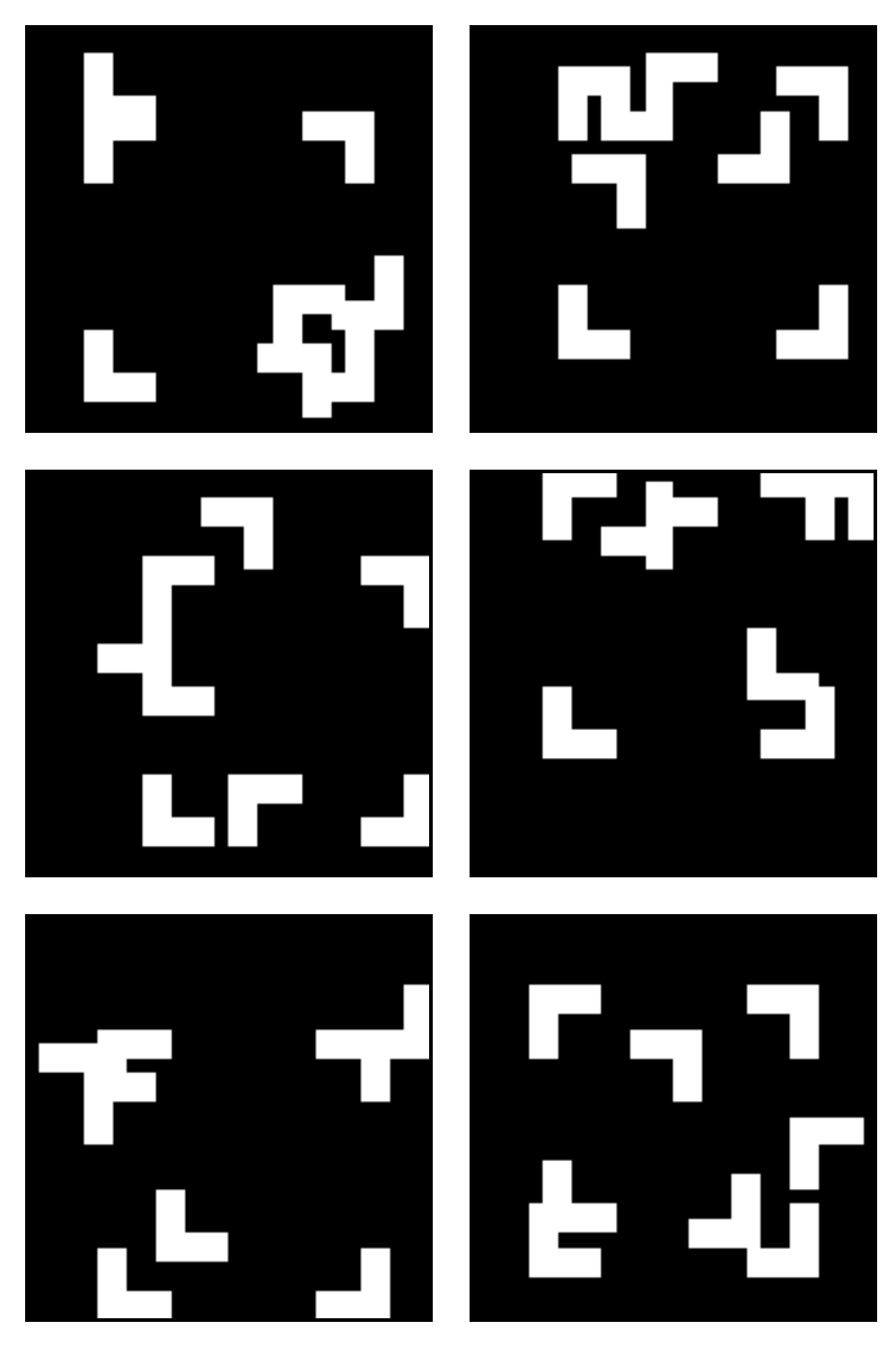}

}\hfill{}\subfloat[]{\includegraphics[width=0.35\textwidth]{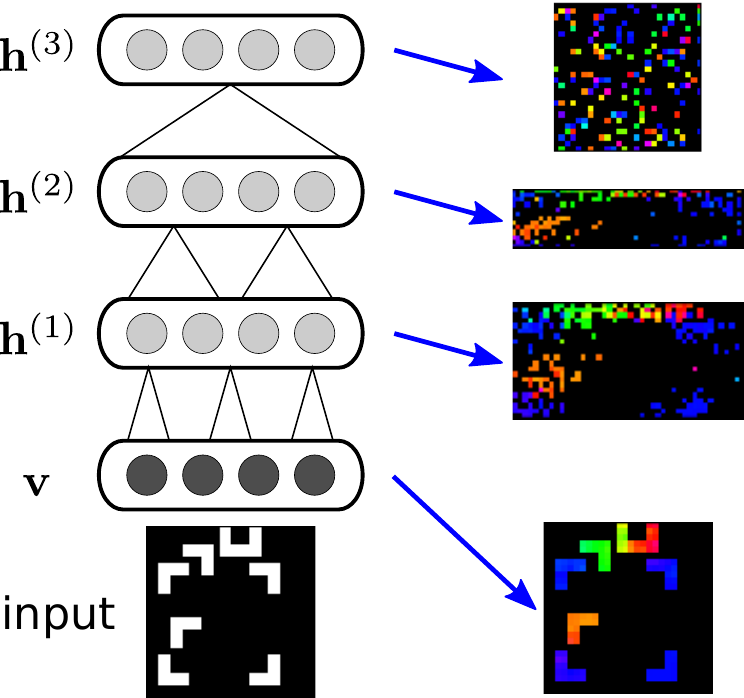}

}

\caption{\textbf{Binding by synchrony in a deep network.} \label{fig:corners_results}
(a) Each image contained four corners arranged in a square shape,
and four randomly positioned corners. (b) The four corners arranged
in a square were usually found to synchronize. The synchronization
of the corresponding hidden units is also clearly visible in the hidden
layers. The receptive field sizes in the first hidden layer were too
small for a hidden unit to `see' more than individual corners. Hence,
the synchronization of the neurons representing the square in the
fist hidden and visible layers was due to feedback from higher layers
(the topmost hidden layer had global connectivity).}
\end{figure}

To examine the effects of synchrony in deeper networks, we trained
a DBM with three hidden layers on another dataset, consisting of binary
images that contained both four `corners' arranged in a square shape
(centered at random positions) and four corners independently drawn
(Figure \ref{fig:corners_results}a). Receptive field sizes in the
first hidden layer were chosen such that the fields would only cover
individual corners, not the whole square arrangements, making it impossible
for the first hidden layer to discover the latter during training.%
\footnote{Note that there was only layer-wise pretraining, no training of the
full DBM, thus first layer representations were not influenced by
higher layers during training either.%
} Receptive field sizes were larger in higher layers, with the topmost
hidden layer being fully connected.

After converting the net to complex values, we found that the four
corners arranged as a square would often synchronize to one phase,
whereas the other, independent corners would assume one or multiple
phases different from the phase of the square (Figure \ref{fig:corners_results}b;
more examples Figure~S1c). Synchronization was also clear in the
hidden layers. 

Again we make several observations. First, because of the restricted
receptive fields, the synchronization of the units representing parts
of the square in the visible layer and first hidden layer was necessarily
due to top-down input from the higher layers. Whether or not a corner
represented by a first layer neuron was part of a larger whole was
made explicit in the synchronous state. Second, this example also
demonstrates that neurons need not synchronize through connected image
regions as was the case in the bars experiment. Lastly, note that,
with or without synchrony, restricted receptive fields and topographic
arrangement of hidden units in intermediate hidden layers make it
possible to roughly identify which units participate in representing
the same image content, by virtue of their position in the layers.
This is no longer possible with the topmost, globally connected layer.
By identifying hidden units in the topmost layer with visible units
of similar phase, however, it becomes possible to establish a connection
between the hidden units and what they are activated by in image space.

\subsection{Reading out object representations via phase\label{sub:Reading-out-object}}

\begin{figure*}[t]
\hspace{1.5cm}\subfloat[]{\includegraphics[height=0.12\textheight]{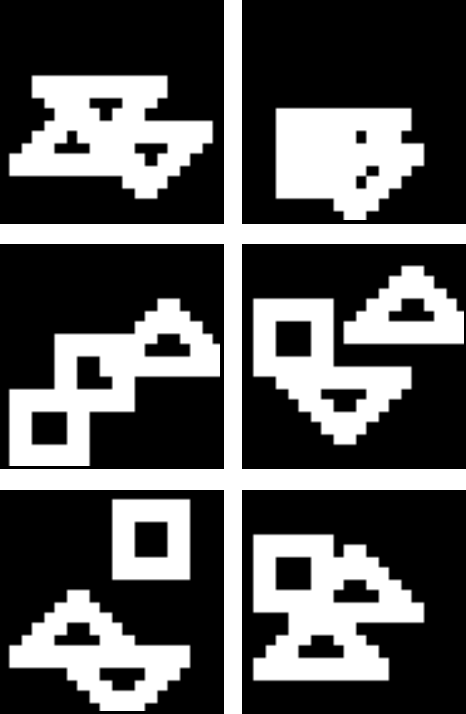}}\qquad{}\subfloat[]{\includegraphics[height=0.12\textheight]{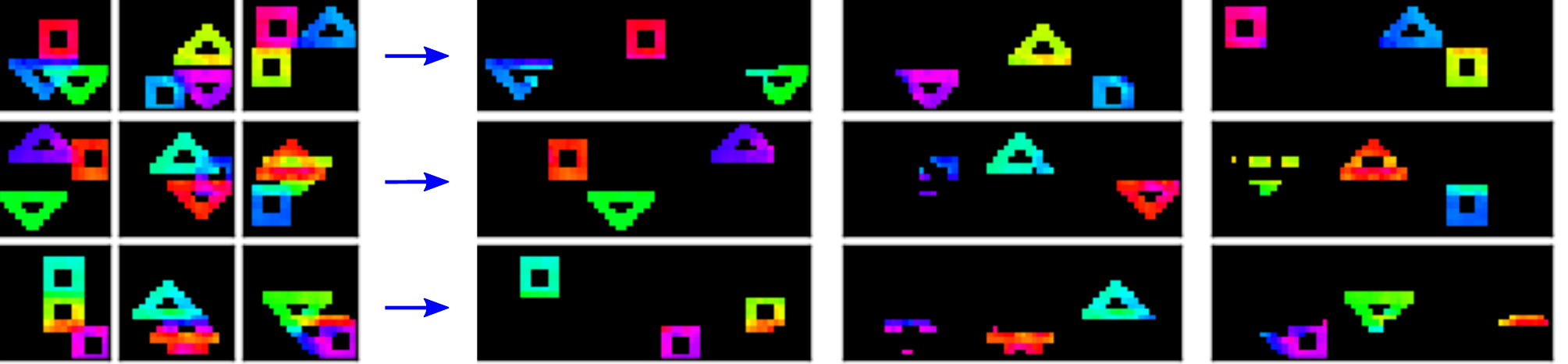}}

\hspace{1.5cm}\subfloat[]{\includegraphics[height=0.12\textheight]{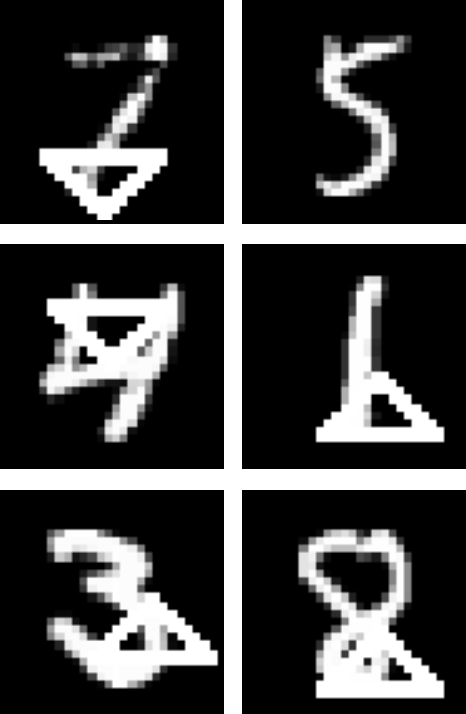}}\qquad{}\subfloat[]{\includegraphics[height=0.12\textheight]{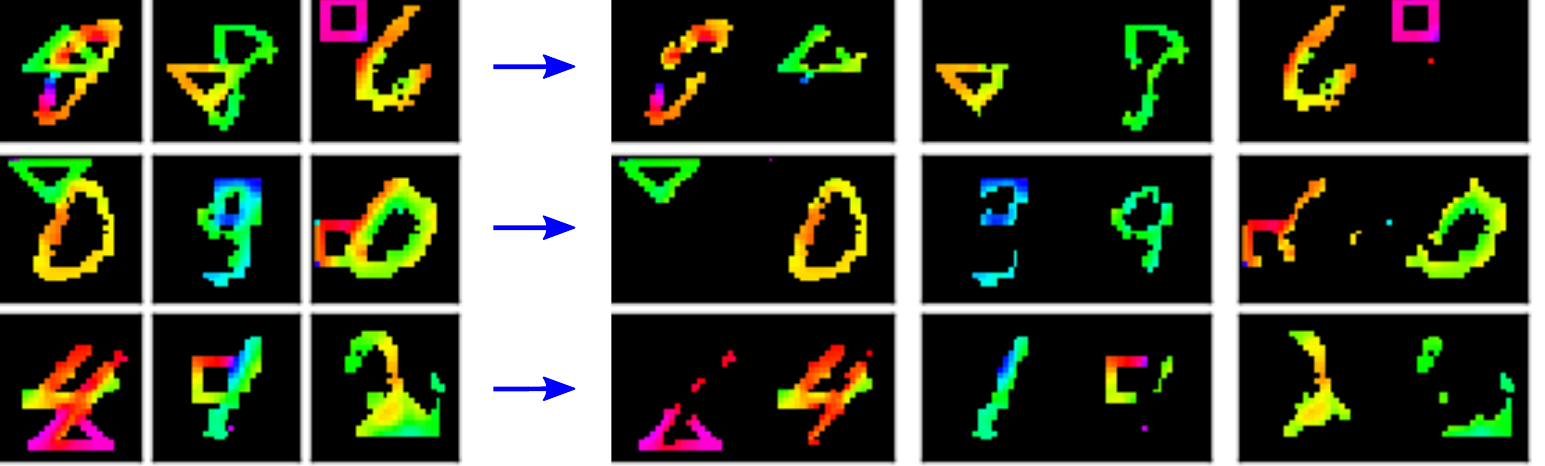}}\caption{\textbf{Simple segmentation from phases.} \label{fig:shapes_MNIST_shape_data}
(a) The 3-shapes set consisted of binary images each containing three
simple geometric shapes (square, triangle, rotated triangle). (b)
Visible states after synchronization (left), and segmented images
(right). (c) For each image in the MNIST+shape dataset, a MNIST digit
and a shape were drawn each with probability 0.8. (d) Analogous to
(b).}
\end{figure*}

With a final set of experiments, we demonstrate that individual synchrony
assemblies can be selected on the basis of their phase, and their
representational content be accessed one group at a time. We trained
on two additional datasets containing multiple simple objects: one
with images of geometric toy shapes (triangles or squares, Figure
\ref{fig:shapes_MNIST_shape_data}a), with three randomly chosen instances
per image, and a dataset where we combined handwritten digits from
the commonly used MNIST dataset with the geometric shapes (Figure
\ref{fig:shapes_MNIST_shape_data}c). As before, we found a tendency
in the complex-valued network to synchronize individual objects in
the image to distinct phases (Figure \ref{fig:shapes_MNIST_shape_data}b,
d, Figure S1d, e). 

After a network was run for a fixed number of steps, for each layer,
units were clustered according to their activity vectors in the complex
plane. For clustering we assumed for simplicity that the number of
objects was known in advance and used k-means, with $k$, the number
of clusters, being set to the number of objects plus one for a general
background. In this fashion, each neuron was assigned to a cluster,
and the assignments could be used to define masks to read out one
representational component at a time.%
\footnote{Alternatively, peaks could be selected in the phase histogram of a
layer and units masked according to distance to the peaks, allowing
for overlapping clusters.%
}

For the visible layer, we thus obtained segmented images as shown
in Figures \ref{fig:shapes_MNIST_shape_data}b, d. Especially for
the modified MNIST images, the segmentations are often noisy. However,
it is noteworthy that segmentations can be obtained at all, given
that the network training involved no notion of segmentation. Moreover,
binding by synchrony is more general than segmentation (in the sense
of assigning labels to pixels), as it applies to \emph{all} neurons
in the network and, in principle, to arbitrarily abstract and non-visual
forms of representation.

\begin{SCfigure*}
\includegraphics[width=0.6\textwidth]{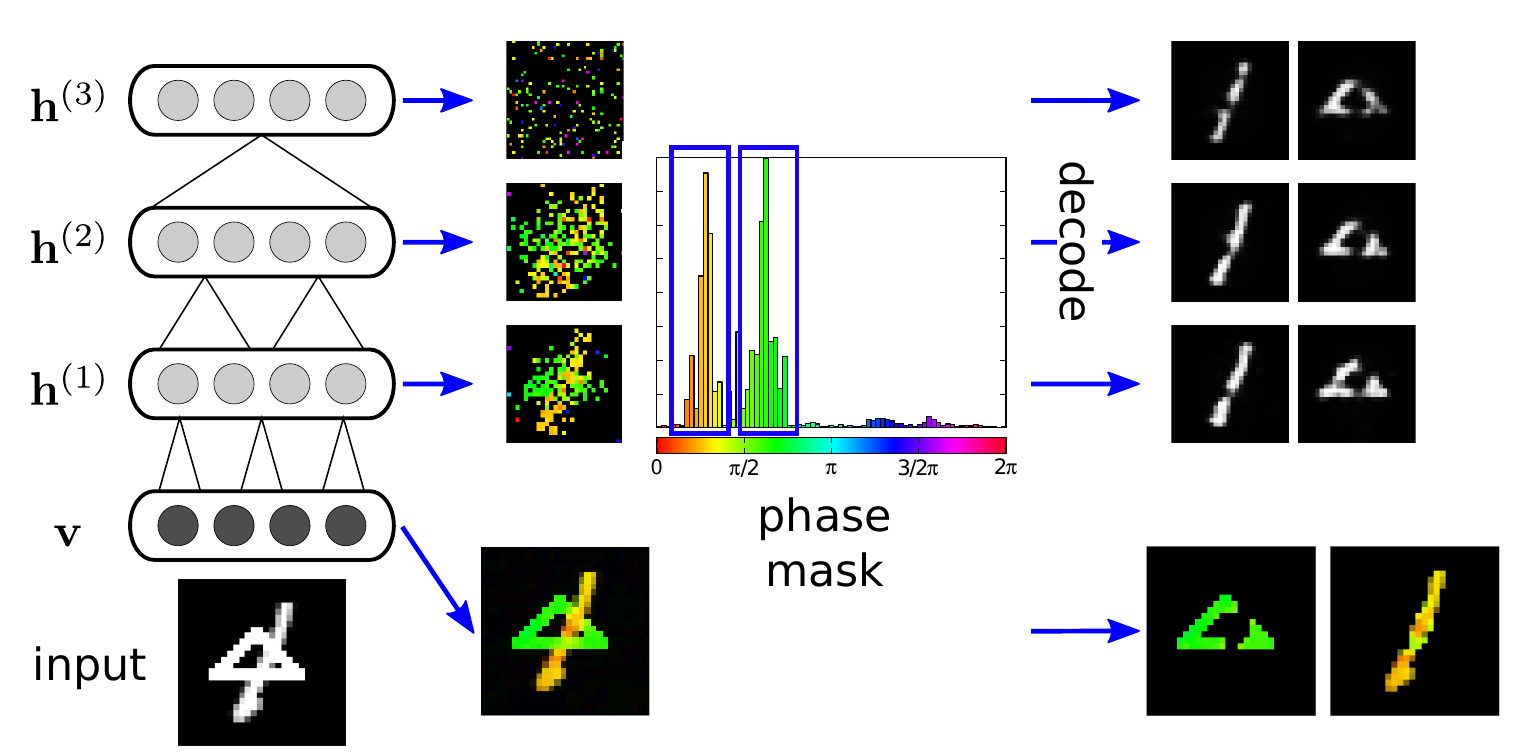}\caption{\textbf{Using phase to access and decode internal object representations.\label{fig:mnist-shape-decode}}
By selecting subsets of neurons according to their phase (e.g.~through
clustering), representations of each object can be read out one by
one (right-hand side). For the hidden layers, plotted are images decoded
from each of the synchronous sub-populations, using a simple decoding
procedure (see main text). }
\end{SCfigure*}

Thus, units can also be selected in the hidden layers according to
phase. The phase-masked representations could, for instance, be used
for classification, one object at a time. We can also decode what
these representations corresponded to in image space. To this end,
we took the masked states for each cluster (treating the other states
as being zero%
\footnote{This only works in a network where units signal the presence of image
content by being on and not by being off, so that setting other units
to zero has the effect of removing image content. This can be achieved
with inductive biases such as sparsity being applied during training,
see the discussion of \citet{honkela_hierarchical_2011}.%
}) and used a simple decoding procedure as described by \citet{reichert_hallucinations_2010,reichert_deep_2012},
performing a single deterministic top-down pass in the network (with
doubled weights) to obtain a reconstructed image. See Figure \ref{fig:mnist-shape-decode}
for an example. Though the images decoded in this fashion are somewhat
noisy, it is apparent that the higher layer units do indeed represent
the same individual objects as the visible layer units that have assumed
the same phase (in cases where objects are separated well).

Earlier, we discussed gating by synchrony as it arises from the effect
that synchrony has directly on network interactions. Selecting explicitly
individual synchrony assemblies for further processing, as done here,
is another potential form of gating by synchrony. In the brain, some
cortical regions, such as in prefrontal cortex, are highly interconnected
with the rest of the cortex and implement functions such as executive
control and working memory that demand flexible usage of capacity-limited
resources according to context and task-demands. Coherence of cortical
activity and synchrony have been suggested to possibly play a causal
role in establishing dynamic routing between these areas \citep[e.g.][]{benchenane_oscillations_2011,miller_cortical_2013}.
Similarly, attentional processing has been hypothesized to emerge
from a dynamically changing, globally coherent state across the cortex
\citep[e.g.][]{duncan_competitive_1997,miller_cortical_2013}. It
is possible that there are dedicated structures in the brain, such
as the pulvinar in the thalamus, that coordinate cross-cortical processing
and cortical rhythms \citep[e.g.][]{shipp_functional_2003,saalmann_pulvinar_2012}.
In our model, one could interpret selecting synchrony assemblies as
prefrontal areas reading out subsets of neuronal populations as demanded
by the task. Through binding by synchrony, such subsets could be defined
dynamically across many different cortical areas (or at least several
layers in a feature hierarchy, in our model).

\section{Discussion\label{sec:Discussion}}

We argue that extending neural networks beyond real-valued units could
allow for richer representations of sensory input. Such an extension
could be motivated by the fact that the brain supports such richer
coding, at least in principle. More specifically, we explored the
notion of neuronal synchrony in deep networks. We motivated the hypothetical
functional roles of synchrony from biological theories, introduced
a formulation based on complex-valued units, showed how the formulation
related to the biological phenomenon, and examined its potential in
simple experiments. Neuronal synchrony could be a versatile mechanism
that supports various functions, from gating or modulating neuronal
interactions to establishing and signaling semantic grouping of neuronal
representations. In the latter case, synchrony realizes a form of
soft constraint on neuronal representations, imposing that the sensory
world should be organized according to distinct perceptual entities
such as objects. Unfortunately, this melding of various functional
roles might make it more difficult to treat the synchrony mechanism
in principled theoretical terms.

The formulation we introduced is in part motivated by it being interpretable
in a biological model. It can be understood as a description of neurons
that fire rhythmically, and/or in relation to a global network rhythm.
Other aspects of spike timing could be functionally relevant,%
\footnote{Consider for instance the tempotron neuron model \citep{gutig_tempotron:_2006},
which learns to recognize spike patterns.%
} but the complex-valued formulation can be seen as a step beyond current,
`static firing rate' neural networks. The formulation also has the
advantage of making it possible to explore synchrony in converted
pretrained real-valued networks (without the addition of the classic
term in Eq.~\ref{eq:input}, the qualitative change of excitation
and inhibition is detrimental to this approach). However, for the
machine learning application, various alternative formulations would
be worthy of exploration (different weights in synchrony and classic
terms, complex-valued weights, etc.).

We presented our simulation results in terms of representative examples.
We did not provide a quantitative analysis, simply because we do not
claim that the current, simple approach would compete with, for example,
a dedicated segmentation algorithm, at this point. In particular,
we found that the conversion from arbitrary real-valued nets did not
consistently lead to favorable results (we provide additional comments in
Appendix \ref{sec:Model-parameters}). Our aim with this paper is
to demonstrate the synchrony concept, how it could be implemented
and what functions it could fulfill, in principle, to the deep learning
community. To find out whether synchrony is useful in real applications,
it is necessary to develop appropriate learning algorithms. We address
learning in the context of related work in the following.

We are aware of a few examples of prior work employing complex-valued
neural networks with the interpretation of neuronal synchrony.%
\footnote{Of interest is also the work of \citet{cadieu_learning_2011}, who
use a complex-valued formulation to separate out motion and form in
a generative model of natural movies. They make no connection to neuronal
synchrony however.%
} \citet{zemel_lending_1995} introduced the `directional unit Boltzmann
machine' (DUBM), an extension of Boltzmann machines to complex weights
and states (on the unit circle). A related approach is used by \citet{mozer_learning_1992}
to model binding by synchrony in vision, performing phase-based segmentation
of simple geometric contours, and by \citet{behrmann_object-based_1998}
to model aspects of object-based attention. The DUBM is a principled
probabilistic framework, within which a complex-valued extension of
Hebbian learning can be derived, with potentially interesting functional
implications and biological interpretations in terms of Spike Timing
Dependant Plasticity \citep{sjostrom_spike-timing_2010}. For our
purposes, the DUBM energy function could be extended to include synchrony
and classic terms (Eq.~\ref{eq:input}), if desired, and to allow
the units to switch off (rather than being constrained to the unit
circle), e.g.~with a `spike and slab' formulation \citep{courville_spike_2011,kivinen_transformation_2011}.
Performing mean-field inference in the resulting model should be qualitatively
similar to running the networks used in our work here.

The original DUBM applications were limited to simple data, shallow
architectures, and supervised training (input phases were provided).
It would be worthwhile to reexamine the approach in the context of
recent deep learning developments; however, training Boltzmann machines
successfully is not straightforward, and it is not clear whether approximate
training methods such as contrastive divergence \citep{hinton_training_2002,hinton_practical_2010}
can be translated to the complex-valued case with success.

\citet{weber_image_2005} briefly describe a complex-valued neural
network for image segmentation, modeling synchrony mediated by lateral
interactions in primary visual cortex, though the results and analysis
presented are perhaps too limited to conclude much about their approach.
A model of binding by synchrony in a multi-layer network is proposed
by \citet{rao_unsupervised_2008} and \citet{rao_objective_2010,rao_effects_2011}.
There, both neuronal dynamics and weight learning are derived from
optimizing an objective function (as in sparse coding, \citealp{olshausen_sparse_1997}).
The resulting formulation is actually similar to ours in several respects,
as is the underlying motivation and analysis. We became aware of this
work after having developed our approach. Our work is complementary
in several regards: our goal is to provide a broader perspective on
how synchrony could be used in neural networks, rather than proposing
one particular model; we performed a different set of experiments
and conceptual analyses (for example, Rao and colleagues do not address
the gating aspect of synchrony); Rao et al.'s approach relied on
the model seeing only individual objects during training, which we
showed to be unnecessary; and lastly, even though they applied synchrony
during learning, the dataset they used for their experiments is, arguably,
even simpler than our datasets. Thus, it remains to be tested whether
their particular model formulation is ideal.

Finally, we are currently also exploring training synchrony networks
with backpropagation. Even a feed-forward network could potentially
benefit from synchrony as the latter could carry information about
sensory input and network state \citep{geman_invariance_2006}, though
complex-valued weights may be necessary for detecting synchrony patterns.
Alternatively, to allow for dynamic binding by synchrony, a network
could be trained as recurrent network with backpropagation through
time \citep{rumelhart_learning_1985}, given appropriate input data
and cost functions. In our experiments, the number of iterations required
was in the order of tens or hundreds, thus making such training challenging.
Again, complex-valued weights could be beneficial in establishing
synchrony assemblies more rapidly. 

\emph{Note: ICLR has an open review format and allows for papers to be updated.
We address some issues raised by the reviewers in Appendix \ref{sec:review}.}

\setcounter{figure}{0} 
\makeatletter 
\renewcommand{\fnum@figure} {Supplementary figure S\thefigure} 
\makeatother

\begin{figure*}[t]
\begin{centering}
\subfloat[]{\includegraphics[width=0.4\textwidth]{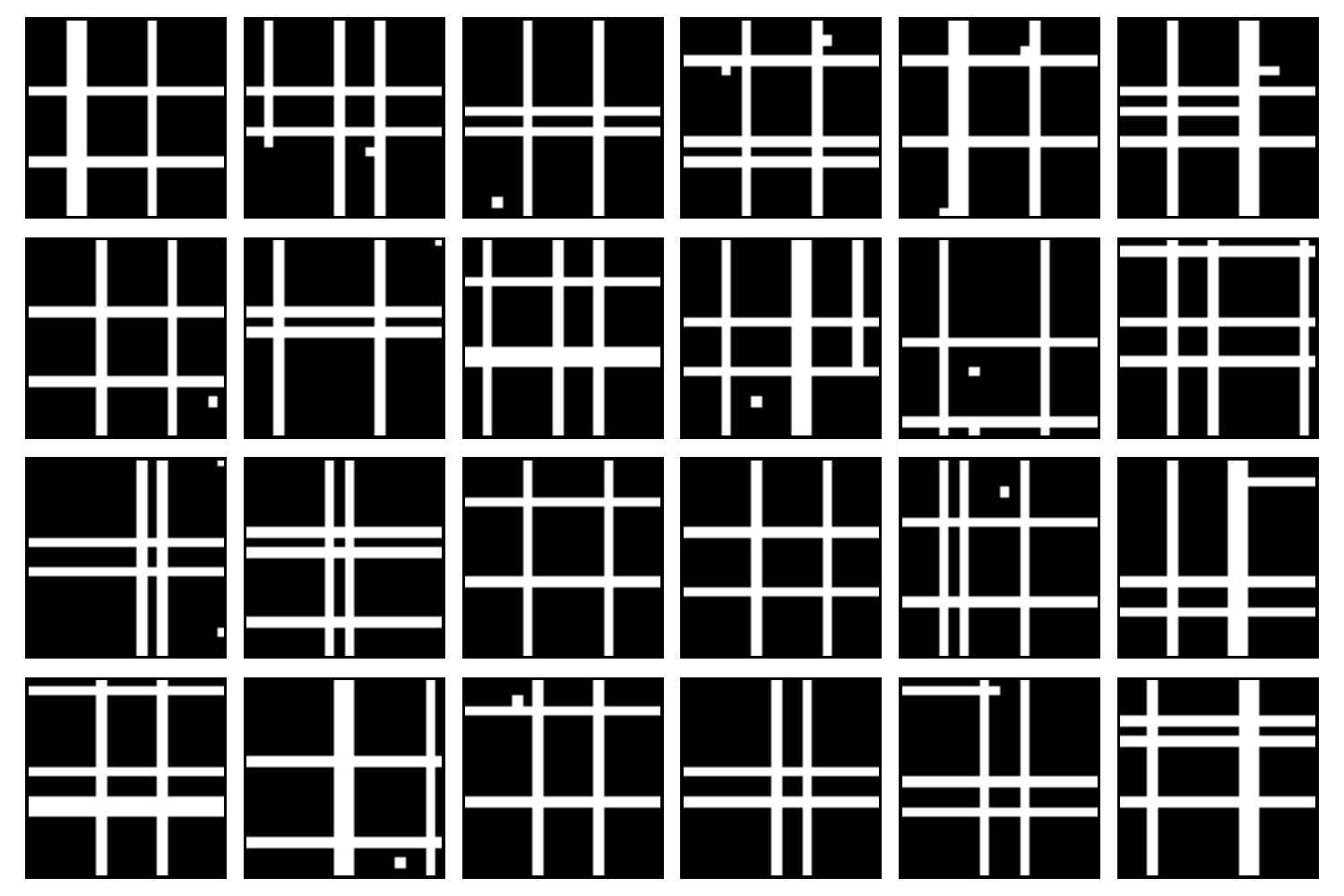}

}\quad{}\subfloat[]{\includegraphics[width=0.4\textwidth]{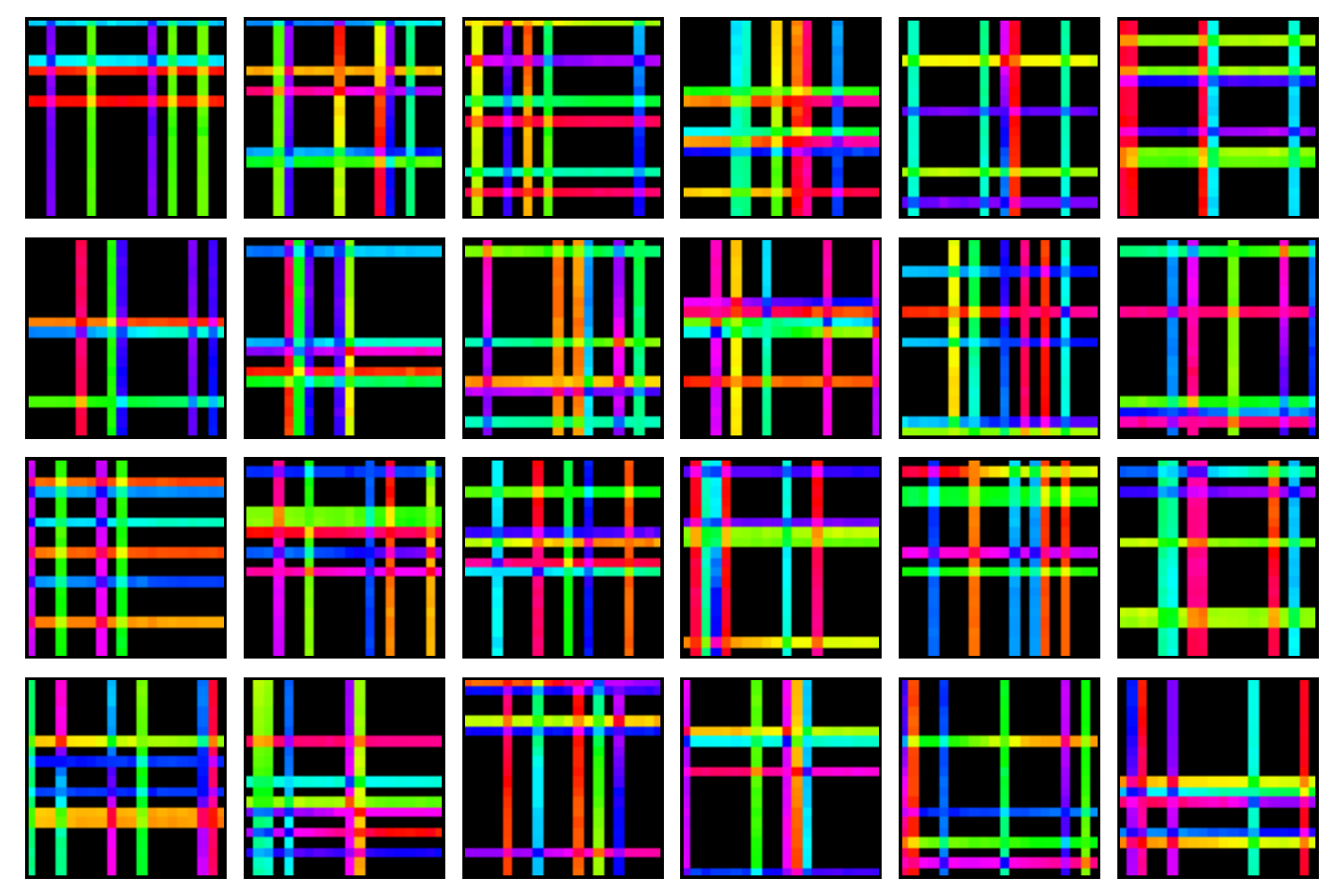}

}
\par\end{centering}

\centering{}\subfloat[]{\includegraphics[width=0.4\textwidth]{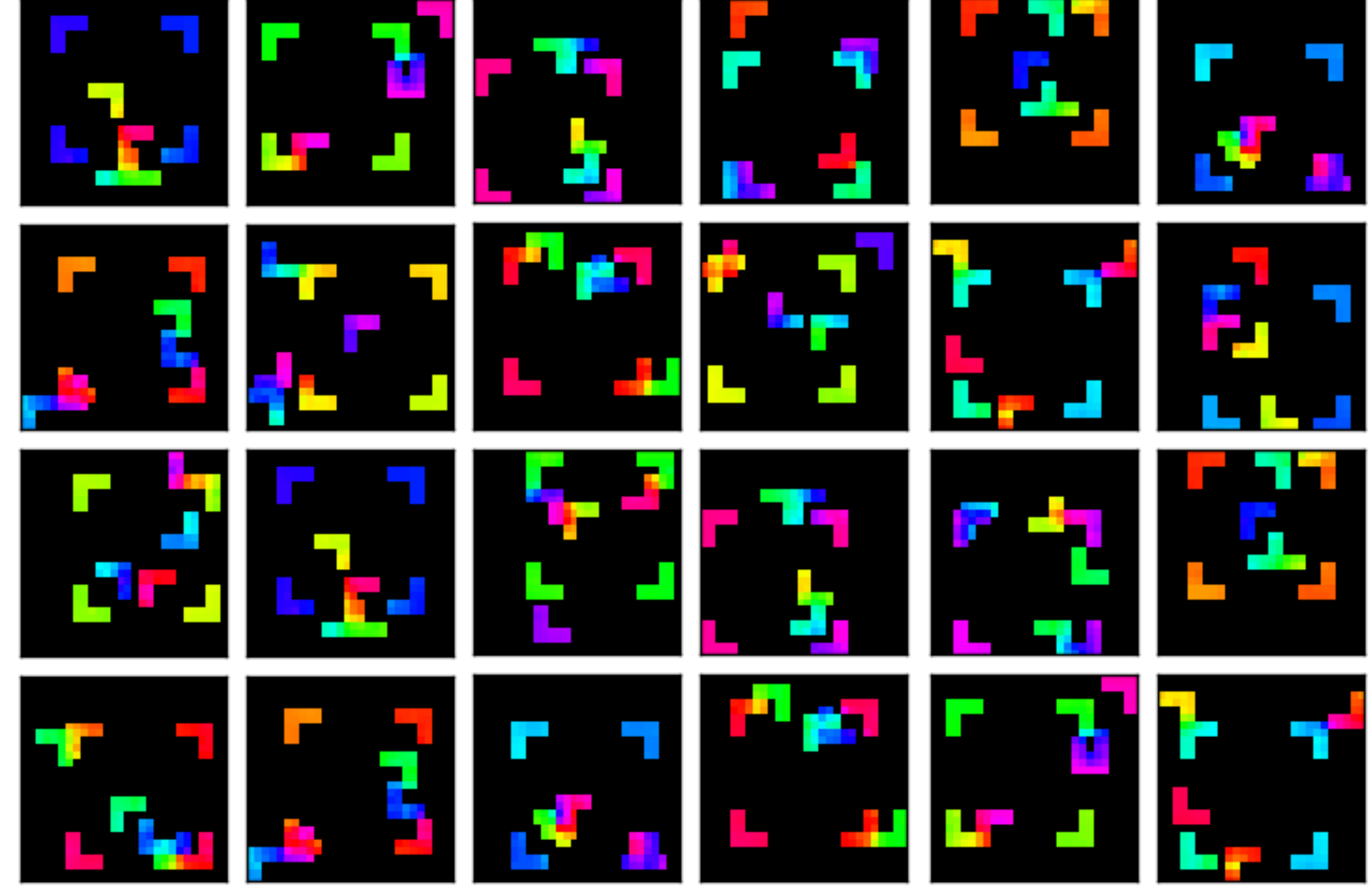}

}\quad{}\subfloat[]{\includegraphics[width=0.2\textwidth]{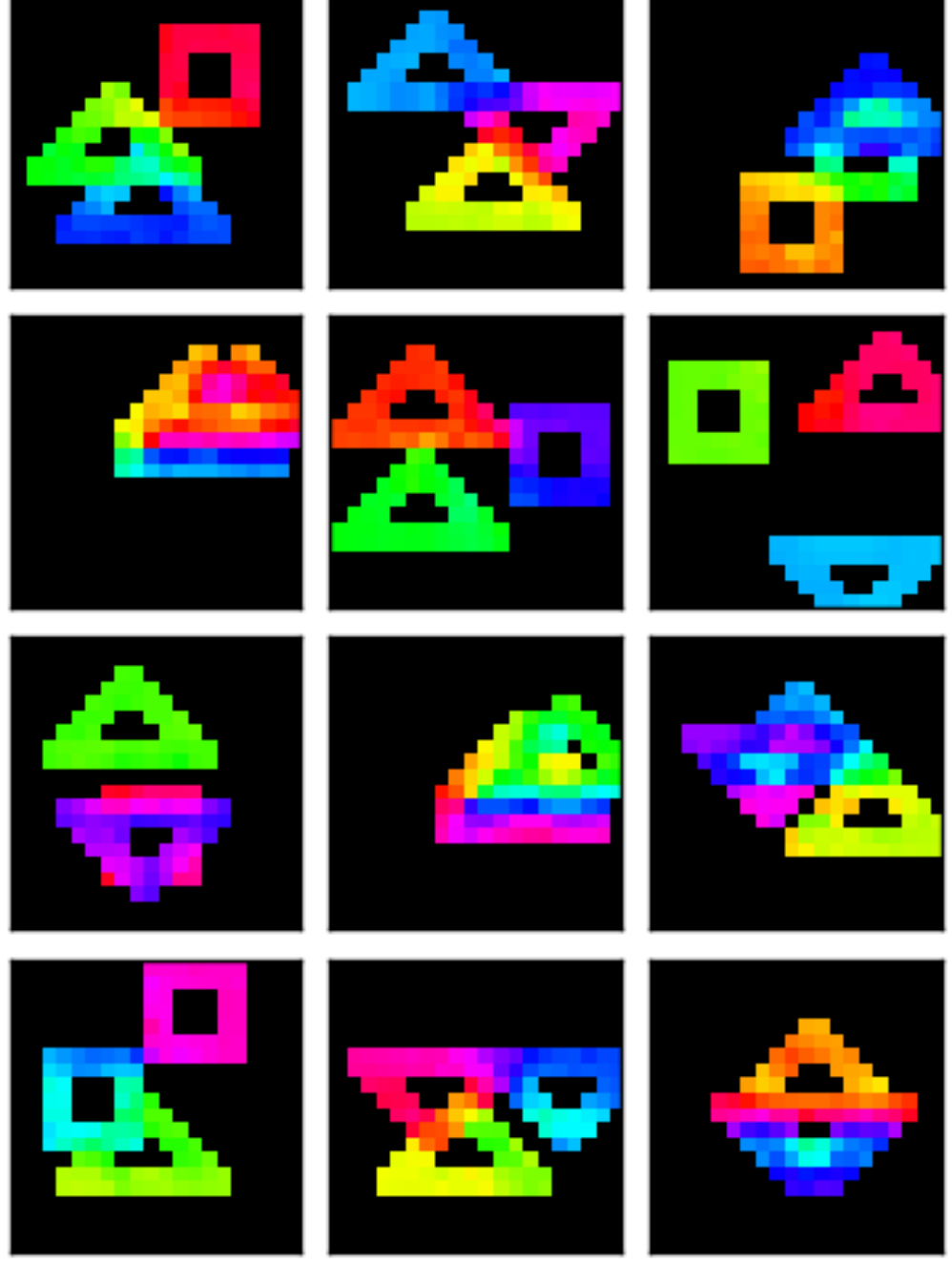}

}\quad{}\subfloat[]{\includegraphics[width=0.2\textwidth]{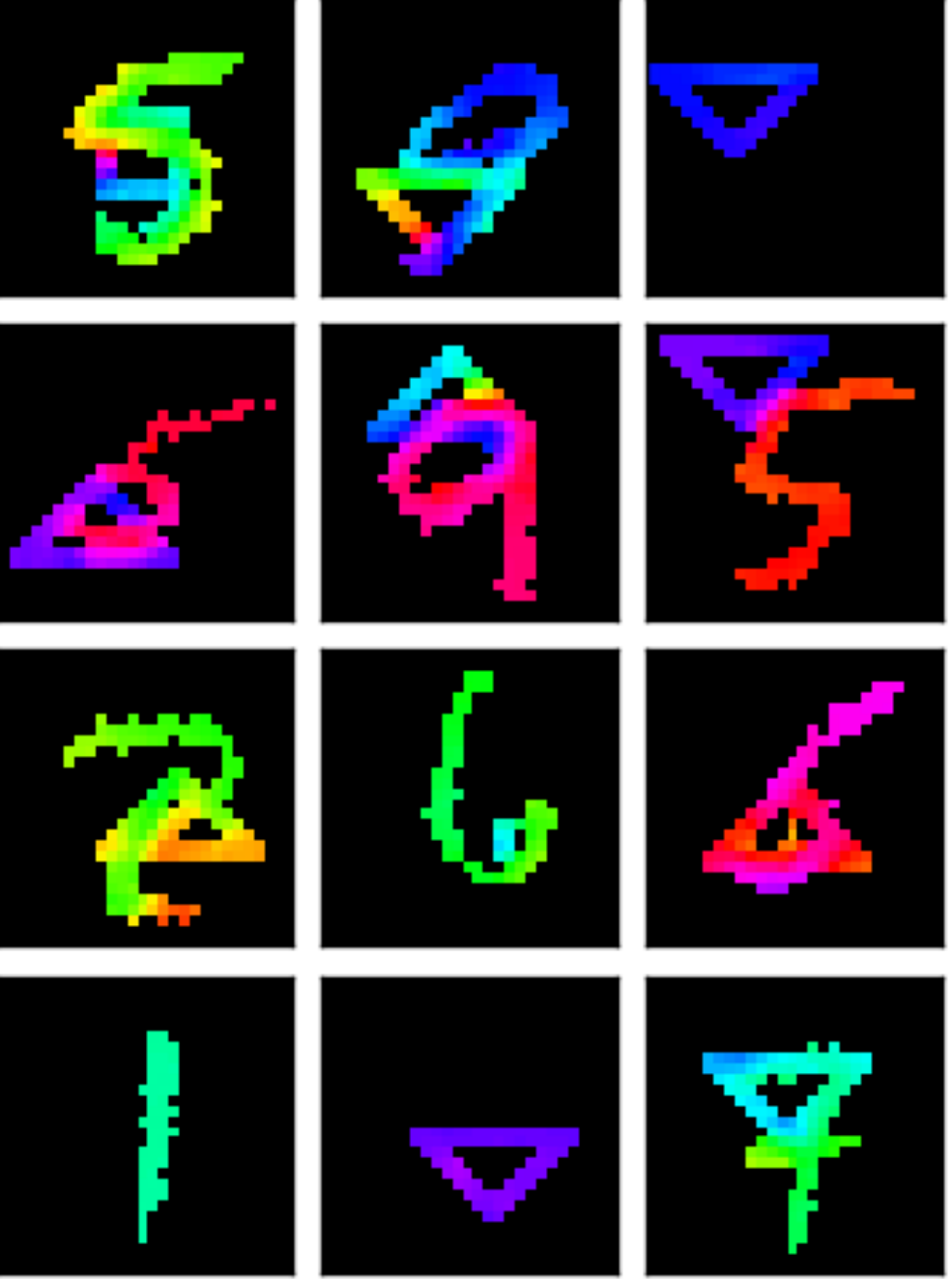}

}\caption{\textbf{Additional results.} \label{fig:Additional-results.-(a)}(a)
Samples generated from a restricted Boltzmann machine trained on the
bars problem. The generated images consist mostly of full-length bars.
The individual receptive fields in the hidden layer were constrained
to image regions of smaller extent than the bars. Thus, bars were
necessarily represented in a distributed fashion. (b) - (e) Additional
examples of synchronized visible units for the various datasets. The
magnitudes of the visible units were set according to the binary input
images (not used in training), the phases were determined by input
from the hidden units. See also supplementary videos (\protect\url{http://arxiv.org/abs/1312.6115}),
and the main text for details.}
\end{figure*}

\subsection*{Acknowledgements}

We thank Nicolas Heess, Christopher K.I. Williams, Michael J. Frank,
David A. Mély, and Ian J. Goodfellow for helpful feedback on earlier
versions of this work. We would also like to thank Elie Bienenstock
and Stuart Geman for insightful discussions which motived this project.
This work was supported by ONR (N000141110743) and NSF early career
award (IIS-1252951). DPR was supported by a fellowship within the
Postdoc-Programme of the German Academic Exchange Service (DAAD).
Additional support was provided by the Robert J. and Nancy D. Carney
Fund for Scientific Innovation, the Brown Institute for Brain Sciences
(BIBS), the Center for Vision Research (CVR) and the Center for Computation
and Visualization (CCV).

\appendix

\section*{Appendix}

\section{Supplementary figures and videos\label{sec:Supplementary-figures}}

Additional outcome examples from the various experiments are shown
in Figure S\ref{fig:Additional-results.-(a)} (as referenced in main
text). We also provide the following supplementary videos on the arXiv
(\url{http://arxiv.org/abs/1312.6115}): a sample being generated
from a model trained on the bars problem (\texttt{bars\_sample\_movie.mp4}),
an example of the synchronization process in the visible and hidden
layers on a bars image (\texttt{bars\_synch\_movie.mp4}), and several
examples of visible layer synchronization for the 3-shapes and MNIST+shape
datasets (\texttt{3shapes\_synch\_movie.mp4} and \texttt{MNIST\_1\_shape\_synch\_movie.mp4},
respectively).

\section{Model and simulation parameters\label{sec:Model-parameters}}

Training was implemented within the Pylearn2 framework of \citet{goodfellow_pylearn2:_2013}.
All networks were trained as real-valued deep Boltzmann machines,
using layer-wise training. Layers were trained with 60 epochs of 1-step
contrastive divergence (\citealp{hinton_training_2002}; learning
rate 0.1, momentum 0.5, weight decay $10^{-4}$; see \citealp{hinton_practical_2010},
for explanation of these training aspects), with the exception of
the model trained on MNIST+shape, where 5-step persistent contrastive
divergence \citep{tieleman_training_2008} was used instead (learning
rate 0.005, with exponential decay factor of $1+1.5\times10^{-5}$).
All datasets had 60,000 training images, and were divided into mini-batches
of size 100. Biases were initialized to -4 to encourage sparse representations
\citep[for reasons discussed by][]{honkela_hierarchical_2011}. Initial
weights were drawn randomly from a uniform distribution with support
$[-0.05,0.05]$.

The number of hidden layers, number of hidden units, and sizes of
the receptive fields were varied from experiment to experiment to
demonstrate various properties of neuronal synchronization in the
networks (after conversion to complex values). The specific numbers
were chosen mostly to be in line with earlier work and not of importance.
In detail, model architectures were as follows: for the bars problem
(Section \ref{sub:Dynamic-binding-of}), input images were $20\times20$,
and the restricted Boltzmann machine had one hidden layer with $14\times14\times3$
units (14 height, 14 width, 3 units per location), and $7\times7$
receptive fields. For the corners dataset (Section \ref{sub:Binding-in-deep}),
input images were $28\times28$, three hidden layers had $22\times22\times2$,
$13\times13\times4$, and 676 units, respectively, and receptive fields
were $7\times7$, $10\times10,$ and $13\times13$ (i.e., global in
the last layer). For the 3-shapes dataset (Section \ref{sub:Reading-out-object}),
input images were $20\times20$, hidden layer dimensions $14\times14\times3$,
$8\times8\times10$, and 676, and receptive fields $7\times7$, $7\times7$,
and $8\times8$ (global). For the MNIST+shape data (also Section \ref{sub:Reading-out-object}),
input images were $28\times28$, hidden layer dimensions $22\times22\times2$,
$13\times13\times4$, and 676, and receptive fields $7\times7$, $10\times10$,
and $13\times13$ (global).

For the synchronization figures, the number of steps to run was chosen
so that synchronization was fairly stable at that point (100 steps
was generally found to be sufficient for all models but the one trained
on MNIST+shape images, where we chose 1000 steps).

Lastly, as mentioned in the main text, we note that the conversion
of pretrained real-valued DBMs did not always lead to models exhibiting
successful synchronization.
Here, successful refers to the ability of the model to separately
synchronize different objects in the input images. Unsuccessful setups
resulted in either all visible units synchronizing to a single phase,
or objects not synchronizing fully, across most of the images in a dataset. 
We found that whether or not a
setup worked depended both on the dataset and the training procedures
used. The presented results are representative 
of well performing networks.

Proper synchronization is an outcome of the right balance
of excitatory and inhibitory connectivity patterns. Further analysis
of how network parameters affect synchronization is the subject of
ongoing work, as is incorporating synchronization during learning
to achieve desired synchronization behavior.

\section{Addressing issues raised by the reviewers\label{sec:review}}

In the following, we summarize parts of the discussion of the ICLR review period, 
paraphrasing the comments of the ICLR reviewers. We expand several
points that were only briefly covered in the main text.

\emph{1.~In the bars experiment, some bars appear to share the same phase. Wouldn't a readout be confused
and judge multiple bars to be the same object?}

This is a very important issue that we are still considering. It is perhaps an issue more generally with the underlying biological theories rather than just our specific approach. As we noted in the main text, some theories pose that a limit on how many discrete objects can be represented in an oscillation cycle, without interference, explains certain capacity limits in cognition. The references we cited \citep{jensen_hippocampal_2005,fell_role_2011} refer to working memory as an example (often 4-7 items; note the number of peaks in Figure \ref{fig:bars_results}c---obviously this needs more quantitative analysis). We would posit that, more generally, analysis of visual scenes requiring the concurrent separation of multiple objects is limited accordingly (one might call this a prediction---or a `postdiction'?---of our model). The question is then, how does the brain cope with this limitation? As usual in the face of perceptual capacity limits, the solution likely would involve attentional mechanisms. Such mechanisms might dynamically change the grouping of sensory inputs depending on task and context, such as whether questions are asked about individual parts and fine detail, or object groups and larger patterns. In the bars example, one might perceive the bars as a single group or texture, or focus on individual bars as capacity allows, perhaps relegating the rest of the image to a general background. 

Dynamically changing phase assignments according to context, through top-down attentional input, should, in principle, be possible within the proposed framework: this is similar to grouping according to parts or wholes with top-down input, as in the experiment of Section \ref{sub:Binding-in-deep}.

\emph{2.What about the overlaps of the bars? These areas seem to be mis- or ambiguously labeled.}

This is more of a problem with the task itself being ill-defined on binary images, where an overlapping pixel cannot really be meaningfully said to belong to either object alone (as there is no occlusion as such). We plan to use (representations of) real-valued images in the future.

\emph{3.~What are the contributions of this paper compared to the work of Rao et al.?}

As we have acknowledged, the work of Rao et al.~is similar in several points (we arrived at our framework and results independently). We make additional contributions. First of all, to clarify the issue of training on multiple objects: in Rao et al.'s work, the training data consisted of a small number of fixed 8$\times$8 pixel images (16 or less images \emph{in total} for a dataset), containing simple patterns (one example has 4 small images with two faces instead). To demonstrate binding by synchrony, two of these patterns are superimposed during test time. We believe that going beyond this extremely constrained task, in particular showing that the binding can work when trained and tested on multiple objects, on multiple datasets including MNIST containing thousands of (if simple) images, is a valid contribution from our side. Our results also provide some insights into the nature of representations in a DBM trained on multiple objects.

Similarly, as far as we can see, Rao et al.~do not discuss the gating aspect at all (Section \ref{sub:The-functional-relevance}), nor the specific issues with excitation and inhibition (Section \ref{sub:Modeling-neuronal-synchrony}) that we pointed out as motivation for using both classic and synchrony terms. Lastly, the following issues are addressed in our experiments only: network behavior on more than two objects; synchronization for objects that are not contiguous in the input images, as well as part vs.~whole effects (Section \ref{sub:Binding-in-deep}); decoding distributed hidden representations according to phase (Section \ref{sub:Reading-out-object}). In particular, it seems to be the case that Rao et al.'s networks had a localist (single object $\leftrightarrow$ single unit) representation in the top hidden layer in the majority of cases.

\emph{4.~The introduction of phase is done in an ad-hoc way, without real justification from probabilistic goals.}

We agree that framing our approach as a proper probabilistic model would be helpful (e.g.~using an extension of the DUBM of \citealp{zemel_lending_1995}, as discussed). At the same time, there is value to presenting the heuristic as is, based on a specific neuronal activation function, to emphasize that this idea could find application in neural networks more generally, not only those with a probabilistic interpretation or Boltzmann machines (that our approach is divorced from any one particular model is another difference when compared to Rao et al.'s work). In particular, we have performed exploratory experiments with networks trained (pretrained as real-valued nets or trained as complex-valued nets) with backpropagation, including (convolutional) feed-forward neural networks, autoencoders, or recurrent networks, as well as a biological model of lateral interactions in V1. A more rigorous mathematical and quantitative analysis is needed in any case.

\emph{5.~How does running the complex-valued network relate to inference in the DBM?}

We essentially use the normal DBM training as a form of pretraining for the final, complex-valued architecture. The resulting neural network is likely not exactly to be interpreted as a probabilistic model. However, if such an interpretation is desired, our understanding is that running the network could be seen as an approximation of inference in a suitably extended DUBM (by adding an off state and a classic term; refer to \citealp{zemel_lending_1995}, for comparison). For our experiments, we used two procedures (with similar outcomes) in analogy to inference in a DBM: either sampling a binary output magnitude from $f()$, or letting $f()$ determine the output magnitude deterministically; the output phase was always set to the phase of the total postsynaptic input. The first procedure is similar to inference in such an extended DUBM, but, rather than sampling from a circular normal distribution on the unit circle when the unit is on, we simply take the mode of that distribution. The second procedure should qualitatively correspond to mean-field inference in an extended DUBM (see Eqs.~9 and 10 in the DUBM paper), using a slightly different output function.

\emph{6.~Do phases assigned to the input change when running for more iterations than what is shown?}

Phase assignments appear to be stable (see the supplementary movies), though we did not analyze this in detail. It should also be noted that the overall network is invariant to absolute phase, so only the relative phases matter.

{\small{ }}
\end{document}